\documentclass[letterpaper, 10 pt, conference]{ieeeconf}  
\IEEEoverridecommandlockouts

\usepackage{cite}
\usepackage{balance}
\usepackage{amsmath,amssymb,amsfonts}
\usepackage{subfigure}
\usepackage{graphicx} 
\usepackage{textcomp}
\usepackage{multirow} 
\usepackage{xcolor}
\usepackage{float}
\usepackage{tabularray}
\usepackage{hyperref}
\usepackage{array}
\usepackage{booktabs}
\usepackage{verbatim} 
\usepackage{makecell} 
\usepackage{booktabs} 
\usepackage{algorithm}
\usepackage{algpseudocode}

\makeatletter

\newcommand{\Rmnum}[1]{\expandafter\@slowromancap\romannumeral #1@}

\makeatother

\def\BibTeX{{\rm B\kern-.05em{\sc i\kern-.025em b}\kern-.08em
    T\kern-.1667em\lower.7ex\hbox{E}\kern-.125emX}}

\title{\LARGE \bf Real-Time 3D Guidewire Reconstruction from Intraoperative DSA Images for Robot-Assisted Endovascular Interventions}

\author{Tianliang Yao$^{1, 2}$, Bingrui Li$^{3}$, Bo Lu$^{4}$, Zhiqiang Pei$^{5}$, Yixuan Yuan$^{2}$, and  Peng Qi$^{1, 6, *}$ 
\thanks{This work has been accepted for publication by IEEE. Copyright may be transferred without notice, after which this version may no longer be accessible.}
\thanks{This work is supported by the National Key Research and Development Program of China under Grant No. 2023YFB4705200, the National Natural Science Foundation of China under Grant No. 62273257, and the Open Project Fund of State Key Laboratory of Cardiovascular Diseases No.2024SKL-TJ002. \emph{(*Corresponding Author: Peng Qi, email: pqi@tongji.edu.cn)}.}
\thanks{$^{1}$Department of Control Science and Engineering, College of Electronic and Information Engineering, and Shanghai Institute of Intelligent Science and Technology, Tongji University, Shanghai 200092, China;}%
\thanks{$^{2}$Department of Electronic Engineering, Faculty of Engineering, The Chinese University of Hong Kong, Hong Kong SAR 999077, China;}%
\thanks{$^{3}$Department of Artificial Intelligence and Computer Science, School of Computer Science, University of Birmingham, Birmingham B15 2TT, United Kingdom;}%
\thanks{$^{4}$Robotics and Microsystems Center, School of Mechanical and Electric Engineering, Soochow University, Suzhou, Jiangsu 215131, China;}%
\thanks{$^{5}$School of Oriental Pan-Vascular Devices Innovation College, University of Shanghai for Science and Technology, Shanghai 200093, China;}%
\thanks{$^{6}$State Key Laboratory of Cardiovascular Diseases and Medical Innovation Center, Shanghai East Hospital, School of Medicine, Tongji University, Shanghai 200092, China.}%
}

\begin{document}

\maketitle

\begin{abstract}
Accurate three-dimensional (3D) reconstruction of guidewire shapes is crucial for precise navigation in robot-assisted endovascular interventions. Conventional 2D Digital Subtraction Angiography (DSA) is limited by the absence of depth information, leading to spatial ambiguities that hinder reliable guidewire shape sensing. This paper introduces a novel multimodal framework for real-time 3D guidewire reconstruction, combining preoperative 3D Computed Tomography Angiography (CTA) with intraoperative 2D DSA images. The method utilizes robust feature extraction to address noise and distortion in 2D DSA data, followed by deformable image registration to align the 2D projections with the 3D CTA model. Subsequently, the inverse projection algorithm reconstructs the 3D guidewire shape, providing real-time, accurate spatial information. This framework significantly enhances spatial awareness for robotic-assisted endovascular procedures, effectively bridging the gap between preoperative planning and intraoperative execution. The system demonstrates notable improvements in real-time processing speed, reconstruction accuracy, and computational efficiency. The proposed method achieves a projection error of 1.76$\pm$0.08 pixels and a length deviation of 2.93$\pm$0.15\%, with a frame rate of 39.3$\pm$1.5 frames per second (FPS). These advancements have the potential to optimize robotic performance and increase the precision of complex endovascular interventions, ultimately contributing to better clinical outcomes.

\end{abstract}

\section{Introduction}

\begin{figure}[t]
\centering
\includegraphics[width=\linewidth]{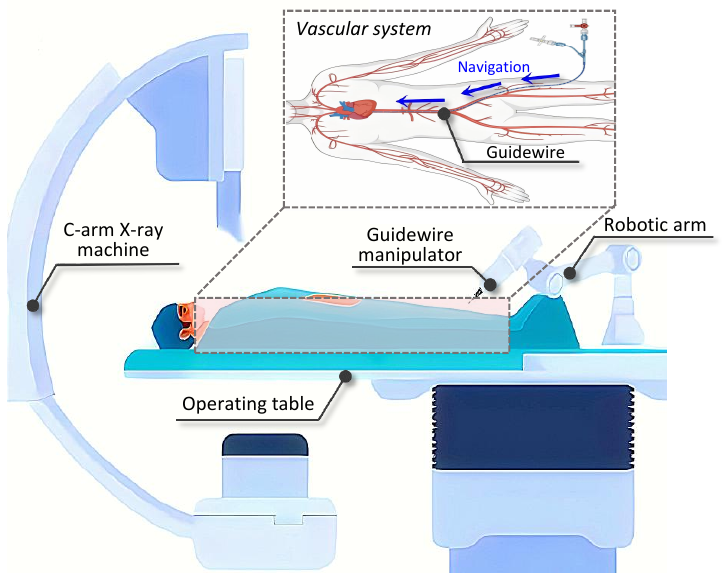}
\caption{The robotic-assisted endovascular procedure is a minimally invasive technique that utilizes interventional guidewires, navigated through blood vessels, to precisely target lesions. During the procedure, the cardiologist relies on real-time 2D Digital Subtraction Angiography (DSA) images provided by a C-arm X-ray machine to operate the guidewire through the intricate vascular network, ensuring accurate navigation to the lesion. The schematic was created using BioRender (\url{https://biorender.com}).}
\label{fig:Illustration}
\end{figure}

Endovascular interventions have evolved into a cornerstone of modern cardiovascular therapeutics \cite{pore2023autonomous, yao2025advancing}, providing minimally invasive treatments for a diverse array of cardiovascular conditions, including arterial stenosis, vascular aneurysms, and valvular insufficiency \cite{nayar2024modeling, wang2024robotic, liu2024magnetic}. A vital element in these interventions is the guidewire, which functions as a flexible, continuum robotic system adept at navigating through the intricate and tortuous vasculature \cite{scarponi2024autonomous, song2024vascularpilot3d, yao2025sim2real}. As depicted in Fig.~\ref{fig:Illustration}, the endovascular procedure involves the minimally invasive manipulation of interventional guidewires through blood vessels to reach targeted lesions. This process critically depends on the precise control of the guidewire to traverse narrow and curved pathways while avoiding damage to adjacent tissues. Accurate navigation of the guidewire to the appropriate vascular bifurcations is indispensable for interventional super-selective procedures \cite{jianu2025splineformer}. This navigation is inherently determined by the guidewire’s spatial morphology, necessitating real-time monitoring of its spatial configuration to ensure precise vessel selection and optimize overall navigation efficacy \cite{li2024model, yao2025sim4endor}.

Existing shape-sensing technologies for continuum robots, such as Fiber Bragg Grating (FBG) and Electromagnetic (EM) sensors, have proven their effectiveness in larger-scale robotic applications \cite{shi2016shape, an2024shape},  including catheter systems and flexible endoscope robots \cite{lu2023autonomous, ourak2021fusion, liu2024desectbot}. However, these methodologies are unsuitable for guidewires due to the challenges posed by their miniature size and the intricacies of integrating such sensors without compromising their flexibility and performance. The small diameter of guidewires, coupled with the necessity to maintain their mechanical properties during navigation, makes it particularly challenging to incorporate traditional embedded sensors, thereby hindering the achievement of precise, real-time shape sensing based on sensors.

Recent advancements in vision-based shape sensing for continuum robotics offer promising solutions to the challenges associated with flexible, thread-like continuum instruments. Ma \textit{et al}. \cite{ma2024shape} introduced a Shape-Guided Configuration-Aware Learning approach for endoscopic-image-based pose estimation, demonstrating significant potential in enhancing real-time tracking and pose estimation of flexible robotic instruments. Furthermore, within the context of the Da Vinci Robotic Surgical System, considerable progress has been made in the 3D shape perception of flexible, thread-like continuum instruments. Several frameworks have showcased significant advancements in this field. For example, Lu \textit{et al}. \cite{lu2020learning} proposed a learning-driven architecture that provides an effective solution for surgical suture tracking through optimized stereo matching and noise-robust feature extraction. Their subsequent model-free approach \cite{lu2021toward} achieved submillimeter reconstruction accuracy in calibrated stereoscopic systems by leveraging adaptive feature point clustering and spatial geometry constraints. Additionally, Joglekar \textit{et al}. \cite{joglekar2023suture} employed Minimum Variation Spline optimization on detected key points, enhancing the accuracy of grasping point estimation for surgical sutures via parametric spline modeling.

While methods employed in the Da Vinci Robotic Surgical System, which utilizes stereo vision to provide 3D depth and spatial information through dual viewpoints, are highly effective, their applicability to endovascular procedures is limited due to fundamental differences in intraoperative imaging capabilities \cite{yao2023enhancing, jianu20233d, ranne2024aiareseg}. In endovascular procedures, as illustrated in Fig.~\ref{fig:Illustration}, the primary imaging modality is real-time 2D DSA from a C-arm X-ray imaging system, typically offering a single viewpoint at any given moment. While robotic manipulators, combined with intraoperative imaging, enable cardiologists to navigate the guidewire with high precision, this constraint severely restricts the ability to perform accurate 3D reconstruction of instruments using traditional stereo-matching methods. In these procedures, the absence of multiple perspectives significantly limits the ability to accurately perceive the 3D shape of flexible guidewires, especially in complex anatomical scenarios such as vascular bifurcations. This limitation renders traditional stereo-matching methods unsuitable for 3D reconstruction in the endovascular domain.

\begin{figure}[t]
    \centering
\includegraphics[width=\linewidth]{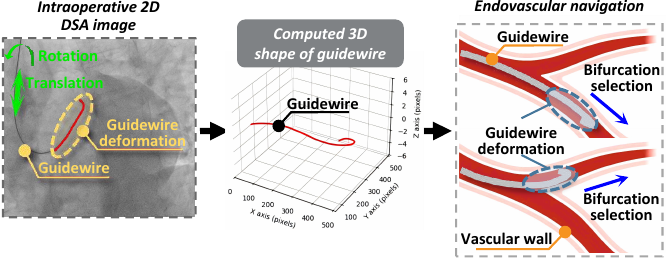}
    \caption{Overview of robotic-assisted endovascular procedures with computational 3D shape reconstruction of the guidewire. During the procedure, real-time visualization is achieved using 2D DSA imaging. However, when selecting the optimal approach at vascular bifurcations, the spatial configuration of the guidewire must be considered. Computational 3D shape reconstruction for guidewire is proposed in this work to provide accurate 3D representations of the guidewire's shape. The schematic was created using BioRender (\url{https://biorender.com}).}
    \label{fig:Overview}
\end{figure}

This paper addresses the limitations of acquiring 3D spatial information of instruments from single-viewpoint, real-time 2D DSA images. A novel methodology is proposed that integrates deformable image registration between preoperative 3D Computed Tomography Angiography (CTA) and intraoperative 2D DSA images. By leveraging the spatial information obtained through registration, this approach enables precise, real-time reconstruction of the guidewire’s 3D spatial configuration, as illustrated in Fig.~\ref{fig:Overview}. This methodology effectively bridges the gap between 2D intraoperative visualization and 3D spatial understanding, facilitating dynamic and accurate guidewire reconstruction in robotic-assisted endovascular procedures.

\begin{figure*}[t]
\centering
\includegraphics[width=0.9\linewidth]{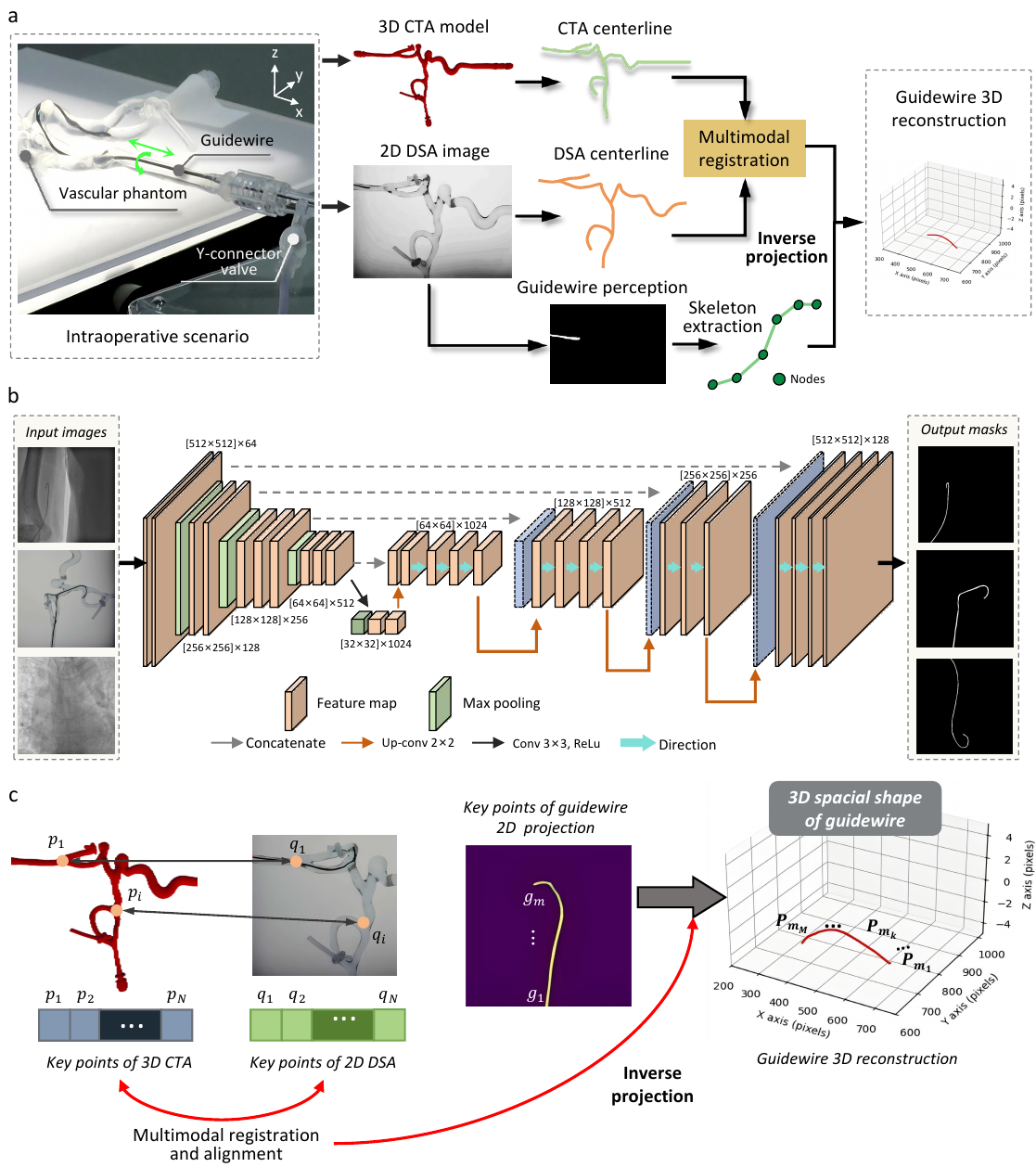}
\caption{Overview of our proposed guidewire 3D reconstruction framework. (a) The pipeline consists of multiple key components: CTA and DSA image processing, centerline extraction, multimodal registration, and 3D guidewire reconstruction. The vascular phantom with structures is used for validation. (b) The guidewire perception module employs UNet with an encoder-decoder structure and skip connections to segment guidewire masks from 2D DSA images. (c) The 3D reconstruction process integrates key points from both imaging modalities: CTA-derived spatial reference points and DSA-based guidewire points are aligned through inverse projection and multimodal registration to achieve accurate 3D guidewire shape reconstruction. The coordinate systems are calibrated in pixels for the 2D projection and spatial dimensions.}
\label{fig:methodology}
\end{figure*}


\section{Methodology}
The proposed framework establishes a diffeomorphic mapping between preoperative 3D vascular structures and intraoperative 2D projections through a cascade of geometrically constrained operators. Given a 3D CTA volume \(\mathcal{V}_{\text{3D}} \subset \mathbb{R}^3\) and 2D DSA sequences \(\{\mathcal{I}_{\text{2D}}^{(t)}\}_{t=1}^T\), the guidewire configuration \(\mathcal{G}_{\text{3D}}(t) \subset \mathbb{R}^3\) is reconstructed in real-time via three fundamental operations: vascular centerline extraction using ridge detection in optimized distance transform fields, Lie group-optimized registration minimizing the reprojection error over \(\text{SE}(3)\) manifold, and depth-aware 3D reconstruction through projective geometry decomposition. The registration phase solves the optimization problem \(T^* = \arg\min_{T \in \text{SE}(3)} \|\pi(T(\mathcal{C}_{\text{3D}})) - \mathcal{C}_{\text{2D}}\|_{\Sigma^{-1}}^2 + \lambda\mathcal{R}(T)\), where \(\pi\) denotes the perspective projection operator and \(\mathcal{R}(T)\) enforces smooth motion priors. Depth estimation leverages the closed-form solution \(d_m = (\mathbf{t}^\top (\mathbf{k}_m \times \mathbf{Rk}_m))/\|\mathbf{k}_m \times \mathbf{Rk}_m\|^2\), derived from epipolar constraints, enabling \(\mathcal{O}(1)\) computational complexity per point. The framework's clinical viability is ensured through uncertainty propagation \(\Sigma_{\mathbf{P}_m} = \mathbf{J}_\pi^\dagger(\sigma_{\text{geo}}^2\mathbf{I} + \sigma_{\text{img}}^2\nabla I\nabla I^\top)\mathbf{J}_\pi^{\dagger\top}\), which quantifies reconstruction precision under imaging noise.

 
\subsection{Preprocessing and Centerline Extraction}  
Let \( \mathcal{M} \subset \Omega \) denote the binary segmentation mask of vasculature from 2D DSA image \( \mathcal{I}_{\text{2D}} \), where \( \Omega \subset \mathbb{Z}^2 \) is the discrete image domain. The morphological skeletonization operator \( \mathcal{S}: \{0,1\}^\Omega \rightarrow \{0,1\}^\Omega \) is defined as: $\mathcal{S}(\mathcal{M}) = \bigcup_{k=0}^K \left( \mathcal{M} \ominus k B \right) \setminus \left( \left( \mathcal{M} \ominus k B \right) \circ B \right)$, where \( B \subset \mathbb{Z}^2 \) is the 3×3 structuring element, \( \ominus \) denotes Minkowski erosion, and \( \circ \) represents the opening operator. The termination index \( K \) satisfies: $K = \max \left\{ k \in \mathbb{N} \ \big| \ \mathcal{M} \ominus k B \neq \emptyset \right\}$, where \( \mathcal{M} \) represents the target image or structure, and \( k \) is the number of iterations of the erosion operation until the result becomes an empty set \cite{onghena2025morphoskel3d}.

The centerline coordinates \( \mathcal{C}_{\text{2D}} = \{\mathbf{q}_j\}_{j=1}^M \) are extracted via connected component analysis: $\mathcal{C}_{\text{2D}} = \left\{ \mathbf{q} \in \Omega \ \big| \ \mathcal{S}(\mathcal{M})(\mathbf{q}) = 1 \ \land \ \sum_{\mathbf{n} \in \mathcal{N}_8(\mathbf{q})} \mathcal{S}(\mathcal{M})(\mathbf{n}) \leq 2 \right\}$ 
where \( \mathcal{N}_8(\mathbf{q}) \) denotes the 8-neighborhood of pixel \( \mathbf{q} \). This ensures the preservation of endpoints and branch points while eliminating spurious skeleton branches. 
 
For 3D CTA volume \( \mathcal{V}_{\text{3D}} \subset \mathbb{Z}^3 \), the centerline extraction utilizes a distance-transform-based approach \cite{song2024iterative}. Let \( \mathcal{M}_{\text{3D}} \subset \mathcal{V}_{\text{3D}} \) be the segmented vessel volume. The Euclidean distance transform \( \mathcal{D}: \mathcal{M}_{\text{3D}} \rightarrow \mathbb{R}^+ \) is computed as $\mathcal{D}(\mathbf{p}) = \min_{\mathbf{q} \in \partial\mathcal{M}_{\text{3D}}}  \|\mathbf{p} - \mathbf{q}\|_2, \quad \forall \mathbf{p} \in \mathcal{M}_{\text{3D}}$. The 3D centerline \( \mathcal{C}_{\text{3D}} = \{\mathbf{p}_i\}_{i=1}^N \) is then obtained through ridge detection $\mathcal{C}_{\text{3D}} = \left\{ \mathbf{p} \in \mathcal{M}_{\text{3D}} \ \big| \ \|\nabla \mathcal{D}(\mathbf{p})\| < \epsilon \ \land \ \lambda_{\text{min}}(\mathbf{H}(\mathcal{D})(\mathbf{p})) < -\delta \right\}$, where \( \mathbf{H}(\mathcal{D}) \) is the Hessian matrix of the distance transform, \( \lambda_{\text{min}} \) denotes its smallest eigenvalue, and \( \epsilon, \delta \) are empirically determined thresholds. 

\subsection{Guidewire Topological Representation} 
The guidewire morphology is formalized as a directed graph $\mathcal{G} = (V, E)$ with node set $V$ and edge set $E$, constructed through the following mathematical formalism. Let the skeletonized guidewire be represented by an ordered coordinate sequence $S = \{v_1, v_2, ..., v_n\}$ where $v_i = (x_i, y_i) \in \mathbb{Z}^2$ denotes the $i$-th pixel position extracted via Zhang-Suen thinning algorithm \cite{zhang2024constraint}. The adjacency matrix $\mathbf{A} \in \{0,1\}^{n \times n}$ encodes structural connectivity: $A_{i,j} = \begin{cases} 
1 & \text{if } \|v_i - v_j\|_2 \leq \sqrt{2} \ \land \ |i - j| = 1 \\
0 & \text{otherwise}
\end{cases}$, where the $\ell_2$-norm condition ensures 8-connected neighborhood topology preservation, and the temporal ordering constraint $|i - j| = 1$ maintains the guidewire insertion sequence. The nodal position matrix $\mathbf{P} \in \mathbb{N}^{n \times 2}$ aggregates spatial coordinates $\mathbf{P} = \begin{bmatrix} (x_i, y_i) \end{bmatrix}_{i=1, \dots, n}$. For enhanced geometric characterization, define the weighted adjacency matrix $\mathbf{W} \in \mathbb{R}^{n \times n}$ incorporating directional derivatives $W_{i,j} = A_{i,j} \cdot \exp\left(-\frac{\|\nabla I(v_i) - \nabla I(v_j)\|^2}{2\sigma_g^2}\right)$, where $\nabla I(v_i) = [\partial_x I(x_i,y_i), \partial_y I(x_i,y_i)]^\top$ denotes the image gradient vector at node $v_i$, and $\sigma_g$ controls feature similarity bandwidth. The graph Laplacian $\mathbf{L} \in \mathbb{R}^{n \times n}$ is subsequently derived as $\mathbf{L} = \mathbf{D} - \mathbf{W}$ with diagonal degree matrix $\mathbf{D}$ where $D_{i,i} = \sum_{j=1}^n W_{i,j}$. This spectral formulation enables Fourier-like analysis of guidewire curvature distribution through eigendecomposition $\mathbf{L} = \mathbf{\Phi} \mathbf{\Lambda} \mathbf{\Phi}^\top$, where $\mathbf{\Lambda}$ contains eigenvalues diagonal matrix and $\mathbf{\Phi}$ the orthonormal eigenvectors.  
 
\subsection{3D-2D Image Registration}  
The registration between 3D coronary centerline \( \mathcal{C}_{\text{3D}} = \{\mathbf{p}_i \in \mathbb{R}^3\}_{i=1}^N \) and 2D vascular projection \( \mathcal{C}_{\text{2D}} = \{\mathbf{q}_j \in \mathbb{R}^2\}_{j=1}^M \) is formulated as a weighted perspective-n-point problem on the Lie manifold \( \text{SE}(3) \). Let \( T \in \text{SE}(3) \) denote the rigid transformation composed of rotation matrix \( R \in \text{SO}(3) \) and translation vector \( \mathbf{t} \in \mathbb{R}^3 \).
 
The perspective projection operator $\pi: \mathbb{R}^3 \rightarrow \mathbb{R}^2$ with intrinsic parameters $K \in \mathbb{R}^{3\times 3}$ maps 3D points to detector coordinates:
 
\begin{equation}
\pi(\mathbf{p}) = \frac{1}{p_z^{(c)}} \begin{bmatrix}
f_x p_x^{(c)} + c_x \\
f_y p_y^{(c)} + c_y 
\end{bmatrix}, \quad \mathbf{p}^{(c)} = R\mathbf{p} + \mathbf{t}
\end{equation}
 
where $(f_x, f_y)$ denote focal lengths and $(c_x, c_y)$ the principal point. The registration energy function combines bidirectional reprojection errors with Mahalanobis weighting:
 
\begin{equation}
\begin{split}
E(T) = & \sum_{i=1}^N \rho\left(\min_{j} \| \pi(T\mathbf{p}_i) - \mathbf{q}_j \|_{\Sigma_j^{-1}}^2 \right) \\
& + \lambda \sum_{j=1}^M \rho\left(\min_{i} \| T^{-1}(\pi^{-1}(\mathbf{q}_j)) - \mathbf{p}_i \|_{\Lambda_i^{-1}}^2 \right)
\end{split}
\end{equation}
 
where $\Sigma_j \in \mathbb{S}_{++}^2$ and $\Lambda_i \in \mathbb{S}_{++}^3$ are covariance matrices encoding 2D/3D detection uncertainties \cite{van20212d}, $\rho(s) = \frac{s}{1+\sqrt{1+s/\delta^2}}$ denotes the pseudo-Huber kernel with cutoff $\delta$, and $\lambda$ balances bidirectional constraints.  
 
The optimization employs manifold-aware Levenberg-Marquardt on Lie algebra $\mathfrak{se}(3)$:
 
\begin{equation}
\boldsymbol{\xi}^{(k+1)} = \boldsymbol{\xi}^{(k)} - \left(\mathbf{J}^\top \mathbf{W} \mathbf{J} + \mu \mathbf{I}\right)^{-1} \mathbf{J}^\top \mathbf{W} \mathbf{r}
\end{equation}
 
where $\boldsymbol{\xi} \in \mathbb{R}^6$ parameterizes $T$ via exponential map $\exp: \mathfrak{se}(3) \rightarrow \text{SE}(3)$, $\mathbf{J} \in \mathbb{R}^{2(N+M)\times 6}$ is the Jacobian of residuals $\mathbf{r}$, and $\mathbf{W} = \text{diag}(\rho'(\|\mathbf{r}_i\|^2))$ provides robust weighting. The damping factor $\mu$ adapts via gain ratio:
 
\begin{equation}
\begin{split}
\mu \leftarrow \mu \cdot \max\left(\frac{1}{3}, 1 - (2\eta - 1)^3\right), \\ \quad \eta = \frac{\|\mathbf{r}^{(k)}\|^2 - \|\mathbf{r}^{(k+1)}\|^2}{\Delta\mathbf{r}^\top (\mu \Delta\boldsymbol{\xi} + \mathbf{J}^\top \mathbf{W} \mathbf{r})}
\end{split}
\end{equation}
 
Convergence is achieved when $\|\boldsymbol{\xi}^{(k+1)} - \boldsymbol{\xi}^{(k)}\| < \epsilon_{\text{abs}}$ and $\|\mathbf{J}^\top \mathbf{W} \mathbf{r}\|_\infty < \epsilon_{\text{rel}} \|\mathbf{r}\|$, guaranteeing first-order optimality while avoiding singular configurations inherent in projective geometry.


\subsection{Guidewire Spatial Configuration Reconstruction}  
The unconstrained 3D reconstruction from 2D projection is formulated through direct geometric inversion. Given the optimal rigid transformation \( T \in \text{SE}(3) \) and camera intrinsic matrix \( K \), the 3D guidewire coordinates \( \mathcal{G}_{\text{3D}} = \{\mathbf{P}_m \in \mathbb{R}^3\}_{m=1}^M \) are recovered via $\mathbf{P}_m = T^{-1}\left( d_m K^{-1} \begin{bmatrix} \mathbf{g}_m \\ 1 \end{bmatrix} \right)$, where \( \mathbf{g}_m \in \mathbb{R}^2 \) denotes 2D guidewire points, and depth \( d_m \) is solved through linear triangulation $d_m = \frac{\| \mathbf{t} \times R^\top K^{-1} \begin{bmatrix} \mathbf{g}_m \\ 1 \end{bmatrix} \|}{\| (R^\top K^{-1} \begin{bmatrix} \mathbf{g}_m \\ 1 \end{bmatrix}) \times (K^{-1} \begin{bmatrix} \mathbf{g}_m \\ 1 \end{bmatrix}) \|}$. The 3D reconstruction can be expressed as a linear mapping $\mathcal{G}_{\text{3D}} = \left\{ 
\begin{bmatrix} 
R^\top & -R^\top \mathbf{t} 
\end{bmatrix} 
\begin{bmatrix} 
d_m K^{-1} \\
\mathbf{g}_m \\
1 
\end{bmatrix}
\right\}_{m=1}^M$

\begin{figure*}[!htbp]
\centering
\includegraphics[width=0.96\linewidth]{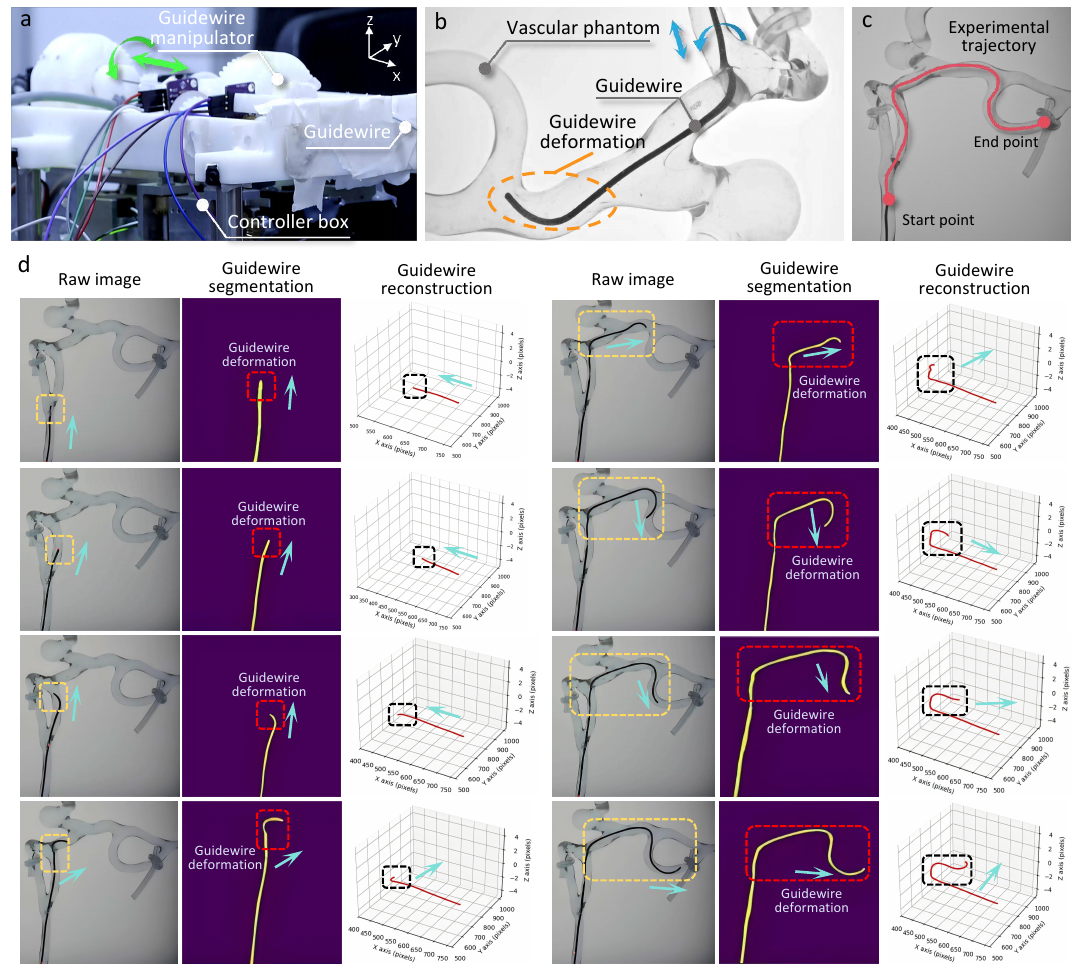}
\caption{Overview of the robotic system and experimental results for 3D spatial reconstruction of a guidewire. (a) The hardware setup of the robotic guidewire manipulation system for endovascular procedures, comprises a custom-designed guidewire manipulator, guidewire, control interface box, and Y-connector valve. (b) Illustration of guidewire deformation during navigation, where the spatial configuration of the guidewire tip determines its correct path selection at vascular bifurcations. (c) Experimental guidewire trajectory from start to endpoint, involving the guidewire manipulator to adjust the guidewire’s shape, with imaging data collected throughout the process. (d) Visualization results of the original DSA images, guidewire segmentation results, and 3D reconstruction of the guidewire's spatial configuration.}
\label{fig:Experimental_Setup}
\end{figure*}


\section{Experiment}

\subsection{Experimal Setup}
The training dataset was acquired from the Department of Cardiology, Zhongshan Hospital, Fudan University, with all sensitive information de-identified. It consists of 799 clinical DSA images, spanning four different sequences. Guidewire annotations were performed at the pixel level by two experienced clinical experts, resulting in an inter-annotator consistency coefficient of 0.94, as confirmed through cross-validation and expert discussion. The images were processed using MicroDicom DICOM Viewer (32-bit) and labeled with 3D Slicer (version 5.6.2).

The testing dataset, consisting of 327 frames from five experiments, was acquired using a vascular intervention robotic system developed by the United Imaging Research Institute. A vascular phantom, also provided by the institute, was used for registration and reconstruction. Both the 3D CTA model and the 2D images were generated using CT scanners and imaging systems from the United Imaging Research Institute of Intelligent Imaging. The experimental setup for the 3D spatial reconstruction of the guidewire is illustrated in Fig.~\ref{fig:Experimental_Setup}(a) depicts the hardware configuration of the robotic system, consisting of a custom-designed guidewire manipulator, the guidewire itself, a control interface box, and a Y-connector valve. Fig.~\ref{fig:Experimental_Setup}(b) highlights the guidewire deformation observed during navigation, which is the central focus of this study. Fig.~\ref{fig:Experimental_Setup}(c) illustrates the navigation task performed during the experiment, showcasing the guidewire’s movement and manipulation task.

The deep learning model was trained on an NVIDIA GeForce RTX\texttrademark~4070 GPU using PyTorch 2.1.2. Data preprocessing leveraged OpenCV 4.10.0.84, resizing 2D DSA image sequences to \(512 \times 512\) and normalizing 3D CTA volume intensities to \([0, 1]\). Voxel coordinates from the 3D CTA were mapped to 2D-pixel space for consistency.

The guidewire segmentation network was trained for 500 epochs with a batch size of 32, employing the Adam optimizer with an initial learning rate of \(1 \times 10^{-4}\). The learning rate decayed by 0.1 every 100 epochs, and a weight decay of \(1 \times 10^{-4}\) was applied to mitigate overfitting.

For image registration, the intrinsic camera matrix \(K\) was defined with focal lengths \(f_x = f_y = 3000\) pixels and principal point \((c_x, c_y) = (256, 256)\) pixels. The rigid transformation initialized with an identity rotation \(R\) and translation \(t = [0, 0, 1000]^T\) in pixel coordinates. The Levenberg-Marquardt algorithm was used, with an initial damping factor \(\mu = 1.0\), up to 100 iterations, and a convergence threshold of \(1 \times 10^{-4}\).

Depth estimation was constrained to \([100, 1000]\) pixels, with geometric uncertainty \(\sigma_{\text{geo}} = 2.0\) pixels and image uncertainty \(\sigma_{\text{img}} = 0.1\). A gradient similarity threshold of \(\sigma_g = 0.85\) ensured reliable depth predictions. Geometric constraints utilized centerline extraction thresholds \(\epsilon = 0.5\) and \(\delta = 1.0\) pixels, with a smoothness weight \(\lambda = 0.1\). Robustness was enhanced using a Pseudo-Huber kernel with \(\delta = 1.0\).

\subsection{Performance Assessment of Guidewire Segmentation}

The segmentation performance of the guidewire is assessed by comparing several deep learning models, namely Fully Convolutional Networks (FCN), DeepLabv3+, and the proposed UNet architecture. The evaluation is carried out based on four key metrics: Dice similarity coefficient, Precision, Recall, and Frames Per Second (FPS), as shown in Table~\ref{tab:seg}.

\begin{table}[htbp]
\centering
\caption{Comparative Analysis of Guidewire Segmentation Performance}
\label{tab:seg}
\begin{tabular}{lcccccc}
\toprule
\textbf{Method} & \textbf{Dice (\%)} & \textbf{Precision (\%)} & \textbf{Recall (\%)} & \textbf{FPS} \\
\midrule
FCN & 82.17 & 78.93 & 85.26 & 112 \\
SegNet & 84.59 & 81.24 & 87.91 & 95 \\
PSPNet & 85.82 & 82.47 & 88.63 & 78 \\
DeepLabv3+ & 86.75 & 83.19 & 89.74 & 89 \\
AttentionUNet & 89.41 & 87.28 & 91.86 & 58 \\
\textbf{UNet} & \textbf{93.13} & \textbf{91.87} & \textbf{94.52} & 63 \\
TransUNet & 92.84 & 90.65 & 95.19 & 46 \\
SwinUNet & 91.58 & 89.36 & 93.81 & 42 \\
\bottomrule
\end{tabular}
\end{table}

The results demonstrate that the proposed UNet architecture delivers superior performance across all evaluation metrics, achieving a Dice score of 93.1\%, Precision of 91.8\%, and Recall of 94.5\%. Despite its higher accuracy, the UNet model operates at a lower FPS (63) compared to FCN (112 FPS) and DeepLabv3+ (89 FPS), which illustrates the trade-off between segmentation accuracy and processing speed.

In contrast, DeepLabv3+ yields a Dice score of 86.7\%, Precision of 83.1\%, and Recall of 89.7\%, showing solid performance but still underperforming relative to the proposed UNet model. FCN, while the fastest model with 112 FPS, exhibits the lowest performance across all metrics, with a Dice score of 82.3\%, Precision of 78.9\%, and Recall of 85.2\%.

\subsection{Evaluation Metrics for 3D Guidewire Reconstruction}
Due to the inherent absence of ground truth data for the 3D guidewire reconstruction problem, a suite of complementary evaluation metrics is employed to rigorously assess the performance of the proposed method. These metrics aim to capture both accuracy and efficiency in the reconstruction process.

The \textit{projection consistency error} serves as a primary measure of alignment between the reconstructed 3D guidewire and its 2D projections. This metric is defined as:

\begin{equation}
E_{proj} = \frac{1}{N}\sum_{i=1}^{N}\|P(X_{3D}) - X_{2D}^i\|_2
\end{equation}

where \( P(\cdot) \) denotes the projection operation from the 3D space to the 2D image planes, \( X_{3D} \) represents the reconstructed 3D guidewire, and \( X_{2D}^i \) corresponds to the 2D segmentation in the \( i \)-th view. The projection consistency error quantifies the accuracy with which the 3D structure aligns with its corresponding 2D projections, providing a direct measure of the fidelity of the reconstructed model.

\begin{table}[htbp]
\centering
\caption{Quantitative Evaluation of 3D Guidewire Reconstruction Performance}
\label{tab:recon}
\begin{tabular}{lcccc}
\toprule
\textbf{Method} & \makecell{\textbf{Projection Error}\\ \textbf{(px)}} & \makecell{\textbf{Length}\textbf{ Dev. (\%)}} & \makecell{\textbf{FPS}} &  \\
\midrule
EPnP \cite{fan2024reinforcement} & 2.84 $\pm$ 0.12 & 5.62 $\pm$ 0.22 & 29.8 $\pm$ 1.3 \\
B-spline \cite{joglekar2023suture} & 2.31 $\pm$ 0.10 & 4.18 $\pm$ 0.18 & 34.6 $\pm$ 1.4  \\
\textbf{Proposed} & \textbf{1.76} $\pm$ \textbf{0.08} & \textbf{2.93} $\pm$ \textbf{0.15} & \textbf{39.3} $\pm$ \textbf{1.5}  \\
\bottomrule
\end{tabular}
\end{table}

The proposed method demonstrates a significant improvement in all key metrics when compared to existing techniques, illustrated in Table.~\ref{tab:recon}. Specifically, the projection error is reduced to 1.76 $\pm$ 0.08 pixels, which represents a substantial decrease in error compared to the EPnP (2.84 $\pm$ 0.12 pixels) and B-spline (2.31 $\pm$ 0.10 pixels) methods. This enhancement indicates a higher level of alignment between the reconstructed 3D guidewire and its 2D projections, a critical factor for real-time intervention applications.

Moreover, the length deviation of the reconstructed guidewire is minimized to 2.93 $\pm$ 0.15\%, demonstrating the method's superior accuracy in representing the actual geometry of the guidewire. This improvement is vital for ensuring the precision of the guidewire's interaction with anatomical structures during procedures.

In terms of computational efficiency, the proposed method achieves the highest frame per second (FPS) rate at 39.3 $\pm$ 1.5, which is notably faster than both EPnP (29.8 $\pm$ 1.3 FPS) and B-spline (34.6 $\pm$ 1.4 FPS). This speed is essential for real-time processing, enabling the method to be effectively applied in dynamic surgical environments where decision-making must occur rapidly.

\subsection{Visualization}

The visualization results validate the efficacy of the proposed method in precisely capturing and representing the guidewire's spatial configuration. As depicted in Fig. \ref{fig:Experimental_Setup}(c), the segmentation pipeline demonstrates superior performance by producing smooth, continuous contours of the guidewire, free from any discontinuities or artifacts. This robust segmentation ensures the accurate delineation of the guidewire's trajectory, which is crucial for high-fidelity 3D reconstruction. The method’s ability to capture subtle deformations of the guidewire is particularly evident during navigation through complex vascular structures. At each stage of the procedure, the system effectively adapts to the dynamic shape changes of the guidewire, providing precise segmentation results even in the presence of overlapping or nearby anatomical structures. This continuous, high-quality segmentation directly contributes to the subsequent 3D reconstruction process. 

Furthermore, the 3D reconstruction, as depicted in Fig. \ref{fig:Experimental_Setup}(c), demonstrates the system's ability to reconstruct the guidewire's spatial configuration with high fidelity in the procedure. The reconstructed shape closely aligns with the actual guidewire, accurately reflecting its form and deformations within the vascular environment.

\section{Discussion}

The results presented in this study demonstrate significant improvements in the 3D reconstruction of guidewire shapes using the proposed multimodal framework. The integration of preoperative 3D CTA data with intraoperative 2D DSA images effectively addresses the challenges posed by the limited depth information inherent in 2D DSA. By leveraging advanced feature extraction, deformable image registration, and a physics-constrained inverse projection algorithm, the method provides accurate and real-time spatial information, which is crucial for robotic-assisted endovascular procedures.

Notably, the proposed framework outperforms traditional methods, such as EPnP and B-spline, in both accuracy and real-time performance. The reduction in projection error (1.76 $\pm$ 0.08 pixels) and length deviation (2.93 $\pm$ 0.15\%) illustrates the robustness of the method in maintaining high accuracy during 3D guidewire reconstruction. Additionally, the system's ability to achieve real-time processing speed (39.3 FPS) emphasizes its potential for clinical use, where speed and accuracy are essential for the success of vascular interventions.

While the results are promising, challenges remain in optimizing the balance between segmentation accuracy and processing speed. For instance, the UNet model demonstrated superior segmentation accuracy but operates at a lower FPS compared to other models, like FCN. Future improvements may focus on enhancing real-time performance without compromising the high accuracy of the proposed method.

Furthermore, to increase the robustness and clinical applicability of the proposed method, future work should involve incorporating more diverse datasets, spanning various patient populations and anatomical conditions, for more comprehensive validation. Expanding the dataset will be crucial to test the system's generalizability and ensure its reliability across a wider range of vascular conditions. Additionally, dynamic factors such as heartbeats and patient movement during procedures should be integrated into the model. Addressing these challenges will further enhance the system's ability to operate effectively in dynamic clinical environments, where real-time adjustments are critical for ensuring the success of endovascular interventions.

A promising direction for future improvements lies in the incorporation of physics-embedded properties \cite{yang2024efficient}. By integrating prior knowledge of the physical properties of the guidewire material and its interaction with surrounding tissues, the model can more accurately simulate the guidewire's behavior within the vascular system. Embedding these physical constraints within the reconstruction framework could reduce errors arising from non-ideal interactions and improve the model's reliability, particularly in complex or dynamic clinical scenarios. The integration of physics-driven optimization algorithms, such as finite element analysis (FEA) or physics-based kinematic models, could provide a more comprehensive understanding of the guidewire's motion and its interactions with the vascular system. The fusion of these physical constraints with deep learning models would likely result in a more efficient and precise real-time system, potentially broadening the clinical applications of robotic-assisted endovascular procedures significantly.

\section{Conclusion and Future Work}

This paper introduces a multimodal framework for 3D guidewire reconstruction, which integrates preoperative 3D CTA and intraoperative 2D DSA images. By addressing the limited depth information inherent in 2D imaging, the proposed method achieves 3D reconstruction through multimodal data fusion, specifically for robotic-assisted endovascular procedures. The framework employs advanced feature extraction and deformable image registration techniques to combine these complementary data sources, ensuring accurate and real-time reconstruction of the guidewire's spatial configuration. Furthermore, an inverse projection algorithm refines the guidewire model, thereby enabling intraoperative 3D reconstruction.

Future work will focus on integrating this framework into a simulation platform, incorporating physical priors to facilitate the creation of digital twins and simulation assets within the domain of robotic-assisted endovascular procedures. Additionally, the adoption of advanced machine learning techniques, such as deep reinforcement learning, could enhance the system's adaptability and accuracy in dynamic clinical environments. Expanding the dataset to include diverse vascular conditions and patient populations will also be crucial for validating the system's robustness across various anatomical scenarios.
\bibliographystyle{ieeetr}
\balance
\bibliography{reference}

\end{document}